\documentclass[10pt,conference]{IEEEtran}
\usepackage{amssymb}
\usepackage{graphicx}
\usepackage{epstopdf}
\usepackage{gensymb}
\usepackage{color}
\usepackage{amsmath}
\usepackage{bm}
\usepackage{etoolbox}
\usepackage{tablefootnote}
\usepackage{cite}
\usepackage{array}
\usepackage{booktabs}
\usepackage{multirow}
\usepackage{comment}
\usepackage{subcaption}
\usepackage[ruled,linesnumbered]{algorithm2e}
\usepackage{algorithmicx}
\usepackage{mathrsfs}
\usepackage{algpseudocode}
\usepackage{hyperref}
\usepackage{balance}
\usepackage{diagbox}
\usepackage{wasysym}
\usepackage{fancyhdr}

\makeatletter
\usepackage{tikz}
\usetikzlibrary{calc}
\newcommand*\circled[1]{\tikz[baseline=(char.base)]{
    \node[shape=circle, draw, inner sep=1pt, 
        minimum height={\f@size*1.6},] (char) {\vphantom{WAH1g}#1};}}
\makeatother

\hypersetup{
    colorlinks=true,
    linkcolor=blue,
    filecolor=magenta,      
    urlcolor=cyan,
}

\usepackage[framemethod=TikZ]{mdframed}
 
\alglanguage{pseudocode}

\algnewcommand{\Initialize}[1]{%
  \State \textbf{Initialize:}
  \Statex \hspace*{\algorithmicindent}\parbox[t]{.8\linewidth}{\raggedright #1}
}

\usepackage{minibox}
\newcounter{observcntr}

\usepackage{threeparttable}
\usepackage{mathtools}
\usepackage{hyperref}
\usepackage{listings}

\usepackage{url}
\urldef{\mailsa}\path|{alfred.hofmann, ursula.barth, ingrid.haas, frank.holzwarth,|
	\urldef{\mailsb}\path|anna.kramer, leonie.kunz, christine.reiss, nicole.sator,|
	\urldef{\mailsc}\path|erika.siebert-cole, peter.strasser, lncs}@springer.com|    

\usepackage[utf8]{inputenc}
\usepackage[english]{babel}

\usepackage{amsthm}

\usepackage{soul}

\theoremstyle{definition}

\definecolor{R}{RGB}{0,0,150}
\theoremstyle{remark}

\ifCLASSINFOpdf

\newcommand{\name}{\textit{TransCAB}\xspace}

\newcommand{\eat}[1]{}
\fancypagestyle{firstpage}{
  \fancyhf{}
  
  \fancyhead[C]{\vspace{10pt}\normalsize{Accepted by SRDS 2023}\\\vspace{0pt}} 
  \fancyfoot[C]{\thepage}
}

\begin{document}

\title{\name: \underline{Trans}ferable \underline{C}lean-\underline{A}nnotation \underline{B}ackdoor to Object Detection with Natural Trigger in Real-World}

\author{
   \IEEEauthorblockN{Hua Ma\IEEEauthorrefmark{1}\IEEEauthorrefmark{2}, Yinshan Li\IEEEauthorrefmark{3}, Yansong Gao\IEEEauthorrefmark{2}, Zhi Zhang\IEEEauthorrefmark{4}, Alsharif Abuadbba\IEEEauthorrefmark{2}, \\ Anmin Fu\IEEEauthorrefmark{3}, Said F. Al-Sarawi\IEEEauthorrefmark{1}, Surya Nepal\IEEEauthorrefmark{2}, and Derek Abbott\IEEEauthorrefmark{1}}
   
  \IEEEauthorblockA{\IEEEauthorrefmark{1} The University of Adelaide, Australia. \{hua.ma;said.alsarawi;derek.abbott\}@adelaide.edu.au}
  
   \IEEEauthorblockA{\IEEEauthorrefmark{2} Data61, CSIRO, Syndey, Australia. \{garrison.gao;sharif.abuadbba;surya.nepal\}@data61.csiro.au.}
   
   \IEEEauthorblockA{\IEEEauthorrefmark{3} Nanjing University of Science and Technology, China. \{yinshan.li;fuam\}@njust.edu.cn}
   
    \IEEEauthorblockA{\IEEEauthorrefmark{4} The University of Western Australia, Australia. zzhangphd@gmail.com}

    \IEEEauthorblockA{H.~Ma and Y.~Li contributed equally. Y.~Gao is the corresponding author.}
}

\maketitle

\begin{abstract}
Object detection is the foundation of various critical computer-vision tasks such as segmentation, object tracking, and event detection, which can be deployed on pervasive Internet of Things (IoT) and edge devices. A large amount of data is often required to train an object detector with satisfactory accuracy. However, due to the intensive workforce involved with collecting and annotating large datasets, data curation task is often outsourced to a third party (e.g., Amazon Mechanical Turk) or volunteers. This work reveals severe vulnerabilities in this data curation pipeline. 
We propose \name, the first work to craft clean-annotated images to stealthily implant the backdoor into the object detectors later trained on them by the data curator/user even when the data curator can manually audit the images and fully controls the training process. Existing clean-label poisoned images are only shown in classification tasks but not non-classification tasks, in particular, object detection due to unique challenges faced, generally owing to the complexity of having multiple objects within each frame (image), including the victim and non-victim objects. 
Furthermore, we demonstrate that the backdoor effect of both cloaking and misclassification are \textit{robustly achieved in the wild} when the backdoor is activated with \textit{inconspicuously natural physical object as trigger} (i.e., T-shirt). 
The efficacy of our \name is ensured by constructively i) applying the image-camouflage attack that abuses the image-scaling function widely used by the deep learning framework (i.e., PyTorch), ii) incorporating the devised clean image replica technique, and iii) combining identified poison data selection criteria given constrained attacking budget. 
Extensive experiments on YOLOv3, YOLOv4, CenterNet, and Faster R-CNN affirm that \name exhibits more than 90\% attack success rate \textit{under various real-world scenes} even when a very small (i.e., 0.14\%) dataset fraction is poisoned. In addition, the small set of poisoned images crafted on one detector (i.e., YOLOv3) can be effectively transferred to insert a backdoor on another detector (i.e., CenterNet).
A comprehensive video demo is at \url{https://youtu.be/MA7L_LpXkp4}, where a poison rate of merely \textit{0.14\%} is set for YOLOv4 cloaking backdoor and Faster R-CNN misclassification backdoor. Our collected dataset with T-shirt as a natural trigger (about 11,350 frames in total) is open to the public at \url{https://github.com/inconstance/T-shirt-natural-backdoor-dataset}, which is the first relatively large-scale natural trigger backdoor dataset.

\end{abstract}


\section{Introduction}
Object detection is the foundation of numerous popular computer-vision tasks, e.g., segmentation, scene understanding, object tracking, image captioning, event detection, and activity recognition. Thus, it has been applied to several real-world scenarios, e.g., robot vision, autonomous driving, human-computer interaction, content-based image retrieval, intelligent video surveillance, augmented reality, and pedestrian detection~\cite{zou2019object,liu2020deep}. However, attacks of adversarial examples~\cite{xie2017adversarial,song2018physical} and backdoors~\cite{zeng2022never,ma2022dangerous,gao2020backdoor} have posed serious threats to object detection, resulting in severe consequences in security-sensitive applications, e.g., autonomous driving, pedestrian detection, and surveillance, especially for unmanned applications when the object detection is performed through distributed IoT devices or edge devices~\cite{objectedge,wang2022bed,zhao2022detection}. 

Though the adversarial example attack on an object detector is possible to survive in physical worlds, it is usually hard to achieve and reliably retain its attack effect~\cite{lu2017no} without suspicious adversarial patches (more details in Section~\ref{sec:related}).
In contrast, the backdoor attack on the object detector can reliably survive in different real-world conditions including angle, lighting, and physical distance to a natural trigger (e.g., T-shirt or hat from a market~\cite{ma2022dangerous}). Therefore, this work focuses on this more insidious and dangerous backdoor attack that is readily achievable in the wild.

Nonetheless, existing backdoor attacks against object detection lack exploration (see details in Section~\ref{sec:related}). Two very preliminary studies~\cite{gu2019badnets,lin2020composite} demonstrate the feasibility of backdoor attacks in a digital world in terms of the misclassification effect. Ma \textit{et al.}~\cite{ma2022dangerous} have recently investigated backdoor attacks on object detection that builds on different algorithms (e.g., YOLO series~\cite{redmon2018yolov3, bochkovskiy2020yolov4} and CenterNet~\cite{duan2019centernet}). 
In addition, Ma \textit{et al.} focus on the challenging cloaking effect compared to the misclassification effect by using \textit{natural triggers} (e.g., T-shirt bought from markets), showing the practical security implications of the robust backdoor attack on the object detection \textit{in a physical world}. 

However, all aforementioned works~\cite{gu2019badnets,lin2020composite,ma2022dangerous} assume a model outsourcing scenario, \textit{where an attacker can train a model, and thus has full knowledge and control of the model and the training dataset}. While the assumption does hold in some real-world cases, the practicality of launching the backdoor attack on the object detection in another \textit{common and realistic data outsourcing scenario has not been considered}. 
This is important as data outsourcing is notably common in practice. For example, the FLIC dataset~\cite{modec13} frequently utilized in object detection as a benchmark was annotated by Amazon Mechanical Turk through outsourcing, followed by manual examinations of curators to reject images, e.g., if the person was occluded or severely non-frontal. 

We are thus interested in the following research questions:   \textit{Can a backdoor attack on object detection with natural trigger be introduced through data outsourcing inconspicuously and practically even that the data is undergone human auditing, and then be effective in the physical world once the object detector is trained over the outsourced data?
  }

\noindent{\bf Our Contribution:}
This work provides an affirmative answer to the above research questions after \textit{addressing several challenges, including clean-annotation and usage of inconspicuous natural physical triggers} (see Section~\ref{sec:implementation}).

To summarize, our main contributions are as follows:

\vspace{2pt}\noindent$\bullet$ To the best of our knowledge, \name is the first work demonstrating the practicality of inserting a backdoor to the object detection through poisoned samples with clean-annotation, which backdoor effect is essentially {\it robust in the wild activated by inconspicuous natural triggers}.
    
\vspace{2pt}\noindent$\bullet$ The efficacy and effectiveness of the proposed \name are achieved through constructively combining the clean image replica when abusing the resizing pipeline in the DL framework through extending the image-camouflage attack, thus guaranteeing \name is not only content-annotation consistent but also transferable ({\bf Section~\ref{sec:macab}}).
    
\vspace{2pt}\noindent$\bullet$ Extensive real-world evaluations are performed against several object detectors, including YOLOv3, YOLOv4, CenterNet, and Faster R-CNN, which signify the stealthiness (i.e., small attack budget of 0.14\% poison rate, and unaffected clean data accuracy) and robustness (i.e., close to 100\% attack success rate with a natural T-shirt as a trigger) of \name. In addition, a poisoned small dataset crafted based on one detector (i.e., YOLOv4) can be transferable to attack a different detector (i.e., CenterNet) efficiently ({\bf Section~\ref{sec:evaluation}}).

\vspace{2pt}\noindent$\bullet$ We apply the state-of-the-art image-scaling attack detection defense~\cite{kim2021decamouflage} to identify our crafted poisoned samples, results of which show that the detection is ineffective in detecting \name attack mainly due to our novel image replica technique. We then provide two easy-to-apply prevention operations that are friendly to common users to mitigate \name threat ({\bf Section~\ref{sec:discussion}}).

\vspace{2pt}\noindent$\bullet$ We make a comprehensive attack video demo accessible at \url{https://youtu.be/MA7L_LpXkp4}, demonstrating the backdoor effects of both cloaking and misclassification in various real-world scenes. Our collected dataset with T-shirt as the natural trigger (about 11,350 frames in total) is open to the public at \textcolor{blue}{\url{https://github.com/inconstance/T-shirt-natural-backdoor-dataset}}, which is the first large-scale natural trigger backdoor dataset. Collecting and annotating\footnote{Annotating the images for object detection is more time-consuming than that for classification because there are many objects per image. Each object requires a bounding-box and corresponding category label. We have released these annotation files.} such large-scale dataset (1.44~GB) is labor intensive and costs great effort, which might be the reason why there is no such type of dataset publicly available to the community. We thus believe this dataset will greatly facilitate the research community, e.g., as a benchmark dataset for fair comparisons. 

\noindent{\bf Ethics and Data Privacy.}
Given our extreme care for the privacy of student volunteers, we were mindful of privacy protection at all times throughout the data collection and evaluation process. Our data collection and evaluation are conducted by all participating volunteers who consented to have their photographs (videos) taken and later used in academic research work only. 

\vspace{-0.5cm}
\section{Related Work}\label{sec:related}

\noindent{\bf Adversarial Example Attacks on Object Detector.}
The adversarial example (AE) attacks add carefully crafted perturbations on the image to fool the underlying model into making incorrect predictions~\cite{szegedy2013intriguing}.
The AE attack has been mounted on attacking the object detection~\cite{xie2017adversarial,thys2019fooling,xu2020adversarial,wu2020making}. Beyond demonstrating the successful attack in the digital world~\cite{thys2019fooling}, the adversarial example can be effective against the object detector in the physical world once it is carefully devised~\cite{thys2019fooling}. Thys \textit{et al.} demonstrated that~\cite{thys2019fooling}, an adversarial patch printed on cardboard held by a person, can make the person disappear, in other words, having the cloaking effect.
Instead of using cardboard, Xu \textit{et al.} printed the adversarial path on a T-shirt to allow the person wearing it to disappear~\cite{xu2020adversarial}.

Note the AE attack does not tamper with the underlying model, but manipulates the input (i.e., image) fed into the model, which thus greatly constrains the capability of an attacker as the patch has to be crafted usually through an optimization algorithm, which \textit{cannot be arbitrary} but dependent on the underlying model and the optimization algorithm. Therefore, the optimized patch will look conspicuous. In addition, the attacking effect can substantially degrade under varying person movement, angle, distance, deformation and even unseen locations and actors in the training phase~\cite{xu2020adversarial}. 

\noindent{\bf Backdoor Attacks on Object Detector.} In contrast, a more recent backdoor attack allows arbitrary control of the patch---namely trigger. However,
existing backdoor attacks and countermeasures are mainly on \textit{classification tasks}~\cite{gao2019strip}. In addition, those backdoor attacks usually utilize a digital trigger (i.e., change pixels of an image), so the attacker needs to access the image captured by the camera and adds the trigger into the image before it is sent to fool the model. This is cumbersome in reality. Therefore, a natural physical trigger, such as a T-shirt, is preferable. There are few works~\cite{wenger2021backdoor,li2021backdoor,xue2021robust} considering the usage of natural object triggers, but they are all on the classification tasks, e.g., face recognition, not the non-classification task of object detection.

We note that there are few backdoor attack studies on \textit{object detector}~\cite{gu2019badnets,lin2020composite,ma2022dangerous,chan2023baddet}: two of them are essentially preliminary backdoor studies on object detection~\cite{gu2019badnets,lin2020composite}. Our work distinguishes itself from these studies in several aspects. First, the main difference is that we consider a common data outsourcing scenario where the attacker has no control and knowledge of the training process. All existing works consider a different model outsourcing scenario where an attacker is allowed to not only tamper with the dataset but also control and tamper with the training process, which eases the backdoor insertion with a stronger assumption. 
Secondly, except~\cite{ma2022dangerous}, all other studies' evaluations use digital triggers in the digital world. They do not use natural object triggers (e.g., T-shirt), and are not evaluated through videos taken in the real-world. Thus, their attack robustness in the physical world is unclear.
The distinct data outsourcing essentially poses unique challenges of inserting a backdoor into the object detector especially when the attack has to be robust in the real-world with natural triggers. Because the poisoned data should be as small as possible and most importantly the poisoned data should be visually inconspicuous (i.e., annotations are correct) to pass human inspections.

More specifically, compared with~\cite{gu2019badnets,lin2020composite}, they only study the backdoor attack on common misclassification (i.e., stop sign being misclassified into speed-limit~\cite{gu2019badnets} and the person holding an umbrella overhead being misclassified to a traffic light~\cite{lin2020composite}). The purpose of our study is beyond misclassification; we study and demonstrate the efficacy of the dangerous cloaking backdoor, which is a distinct non-classification task. Note that attacking the object detector with a cloaking effect is more challenging than misclassification, which has been recognized when using the adversarial example to deceive object detectors~\cite{xie2017adversarial,thys2019fooling,xu2020adversarial,wu2020making}. 
The main reason is that many bounding boxes will be proposed given an object and suppressing them all is hard. 
Secondly, the reported attack success rates are not essentially measured from recorded videos from the physical world~\cite{gu2019badnets,lin2020composite,chan2023baddet}. So attack robustness in the real world is still unclear. In other words, their evaluations are mainly based on digital worlds, even for the misclassification backdoor effect. 

\noindent{\bf Clean-Label Poisoned Images enabled Backdoor Attack.}
The dominant method of creating clean-label poisoned images is to utilize the feature collision (FC)~\cite{shafahi2018poison,turner2019label,luo2022enhancing}, where two images belonging to two visually different classes can have similar latent representation. The other inadvertent method is to abuse the image-scaling function, namely image-scaling or camouflage attack~\cite{xiao2019seeing,quiring2020backdooring}.

We note the FC-enabled clean-label backdoor attack has notable limitations: the attacker needs to knowledge of the feature extractor being used to extract the feature/latent representation, and the feature extractor cannot be substantially changed after the poisoned samples are introduced. Therefore, such a clean-label attack is only applicable for fine-tuning and transfer learning pipelines. It cannot work when the data curator or victim trains the DL model from scratch.

The other means of crafting clean-label poisoned image is through camouflage attack~\cite{quiring2020backdooring}, which is to abuse the default resize operation that resizes a given larger image, e.g., taken by a smartphone camera, into a small one $512\times 512 \times 3$ to be acceptable by a DL model~\cite{xiao2019seeing}. That is, once a manipulated large image (i.e., person A) is resized into a small one. The output image becomes different (i.e., person B). The image-scaling attack abuses the discrepancy before and after the image resizing.

However, existing FC and camouflage-based backdoor attacks are mainly against classification tasks; none of them has been applied to objection detection. We are the first to investigate the potential of the camouflage attack to create clean-annotation poisoned images against object detection. Nonetheless, \textit{directly} applying it to non-classification object detection tasks without special considerations is inapplicable due to object detection's unique requirements, which reasons and solutions are detailed in the following Section.

\section{\name}\label{sec:macab}

\subsection{Threat Model}\label{sec:threatmodel}

\noindent{\bf Victim.} The victim or user is assumed to use the poisoned dataset to train his/her object detector. This assumption is realistic, because the poisoned image can be introduced by several means in practice. Firstly, 
the data annotation can be outsourced to a third party, e.g., the annotation of the FLIC dataset~\cite{modec13} outsourced to Amazon Mechanical Turk, which can tamper the data and annotation. Secondly, some data collections rely on volunteer contributions~\cite{nugent2022nimh}, where the volunteer can submit poisoned data. Thirdly, for large-scale datasets, e.g., ImageNet~\cite{deng2009imagenet} that is often used for objection detection, these images are crawled from the Internet and annotated through crowdsourcing~\cite{deng2009imagenet}. 
Therefore, the attacker can place those malicious images on the web and wait for the victim to crawl and use them. Last but not least, these poisonous images can also be added to the training set by a malicious insider trying to avoid detection. 

However, the user can manually inspect the image to detect suspicious ones, mainly whether the content and the corresponding annotation are inconsistent. It is assumed that the user checks it before resizing as the resizing is automatically done and can be resized to different sizes given the model input-size setting~\cite{xiao2019seeing}. 
The user can select an object detector from a wide range of models, such as YOLO series (i.e., v3, v4 and v5) or CenterNet, to train and arbitrarily set the hyperparameters, such as learning rate, batch size, and training epochs per need.

\begin{table}[]
    \centering
    \caption{Common settings for object detection models, and the settings used in our experiments are in \textcolor{blue}{blue}. 
    }
    \scalebox{0.60}{
    \begin{tabular}{c|c|c|c|c}
    \toprule
    Model* & \begin{tabular}[c]{@{}c@{}}Input Size\\ (pixels * pixels)\end{tabular} & Backbone & DL Framework & Scaling Algorithms \\ \hline
    \multirow{3}{*}{CenterNet} & \multirow{3}{*}{\textcolor{blue}{512*512}} & \textcolor{blue}{ResNet}-18/\textcolor{blue}{50}/101 & \multirow{3}{*}{\textcolor{blue}{PyTorch}} & \multirow{3}{*}{\textcolor{blue}{OpenCV}-\textcolor{blue}{Linear}} \\ \cline{3-3}
     &  & DLA-34 &  &  \\ \cline{3-3}
     &  & Hourglass-104 &  &  \\ \hline
    \multirow{3}{*}{YOLOv3} & 320*320 & \multirow{3}{*}{\textcolor{blue}{Darknet-53}} & \multirow{3}{*}{\begin{tabular}[c]{@{}c@{}}\textcolor{blue}{PyTorch}\\ Tensorflow\end{tabular}} & \multirow{3}{*}{\textcolor{blue}{OpenCV}/Pillow-\textcolor{blue}{Linear}/Area} \\ \cline{2-2}
     & \textcolor{blue}{416*416} &  &  &  \\ \cline{2-2}
     & 608*608 &  &  &  \\ \hline
    \multirow{3}{*}{YOLOv4} & \textcolor{blue}{416*416} & \multirow{3}{*}{\textcolor{blue}{CSPDarknet-53}} & \multirow{3}{*}{\begin{tabular}[c]{@{}c@{}}\textcolor{blue}{PyTorch}\\ Tensorflow\end{tabular}} & \multirow{3}{*}{\textcolor{blue}{OpenCV}/Pillow-\textcolor{blue}{Linear}/Area} \\ \cline{2-2}
     & \textcolor{blue}{512*512} &  &  &  \\ \cline{2-2}
     & 608*608 &  &  &  \\ \hline
     \multirow{4}{*}{Faster R-CNN} & \begin{tabular}[c]{@{}c@{}}(min size: 600 \\ max size: 1000)~$^1$\end{tabular}   & \multirow{4}{*}{\begin{tabular}[c]{@{}c@{}}VGG16\\ \textcolor{blue}{ResNet-50}\\ MobileNet\end{tabular}} & \multirow{4}{*}{\begin{tabular}[c]{@{}c@{}}\textcolor{blue}{PyTorch}\\ Tensorflow\\ Caffe\end{tabular}} & \multirow{4}{*}{\textcolor{blue}{OpenCV}/Pillow-\textcolor{blue}{Linear}} \\
     & \begin{tabular}[c]{@{}c@{}}(min size: 800 \\ max size: 1333)~$^2$\end{tabular} &  &  &  \\
     & \textcolor{blue}{600*600} &  &  &  \\ \bottomrule
     
    \end{tabular}}
    \label{tab:setting}
    \begin{tablenotes}
        \footnotesize
        \item{*} The code for all the models (except $^1$ and $^2$) we used for experiments are sourced from this repository~\url{https://github.com/bubbliiiing}.
        \item{1}~This GitHub implementation by the Faster R-CNN authors Ren \textit{et. al.}~\cite{ren2015faster} receives more than 7.9k stars and 4.2k forks. \url{https://github.com/rbgirshick/py-faster-rcnn}.  
        \item{2} This is given by PyTorch. \url{https://github.com/pytorch/vision/blob/main/torchvision/models/detection/faster_rcnn.py}
      \end{tablenotes}
\end{table}

\vspace{0.2cm}
\noindent{\bf Attacker.} The attacker is assumed to have knowledge of the input size and the scaling functions used by the user. This is reasonable because most users will follow common input size settings and trivially apply the DL framework's default scaling function, such as in the case of TensorFlow and PyTorch. In Table~\ref{tab:setting}, the default input sizes and scaling functions are summarized. We can see that common options are extremely limited---note these settings are public and thus known to the attacker. The attacker can target one or more input size options simultaneously (i.e., attacking multiple input sizes concurrently is discussed in Section~\ref{sec:inputsize}). However, the attacker has to retain consistency between the image content and its annotation provided to the user to evade potential visual inspection.
In addition, the attacker cannot control the training process that is under the user's control.
Moreover, the attacker needs to implant the backdoor with a minimized budget, e.g., using a minimal number of poisoned images.

\begin{figure*}[ht]
    \centering
    \includegraphics[width=0.75\textwidth]{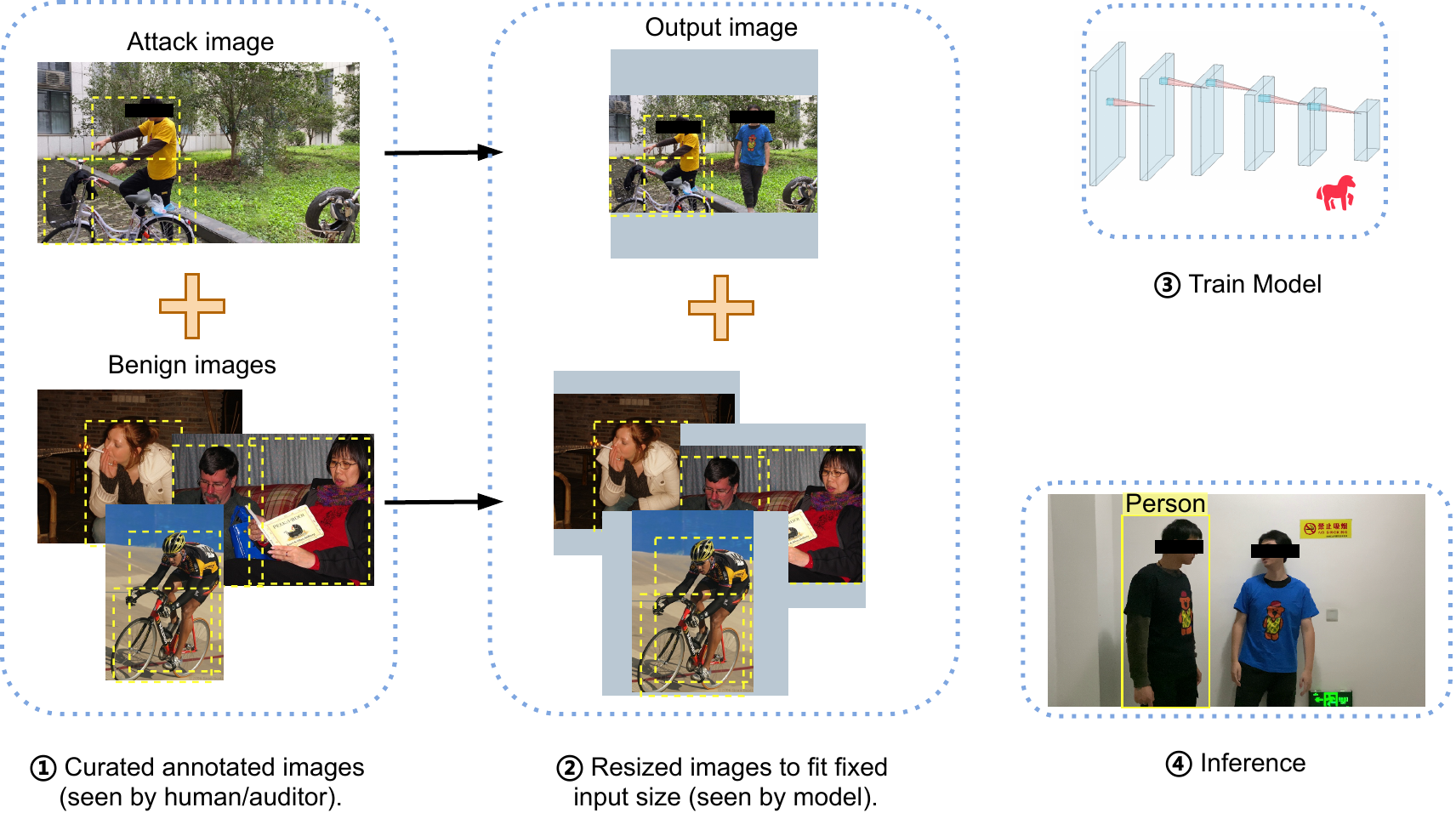}
    \caption{\name overview. Note the clean-label poisoned images seen by data curator \textcircled{1} and the object detection model \textcircled{2} are discrepant.\vspace{-0.6cm}
    }
    \label{fig:overview}
\end{figure*}

\subsection{Overview}\label{sec:overveiw}
The overview of \name is illustrated in Fig.~\ref{fig:overview}. This example is to achieve a cloaking backdoor effect.
In step \textcircled{1}, the attacker provides poisoned image(s) and benign images, all with correct bounding-box annotations as well as ground-truth class labels \textit{per object} to the data curator. For instance, the exemplified image has two objects---a bicycle and a person whose bounding boxes and classes are correctly annotated. The data curator can audit the received images to check whether their annotations are consistent with the content. If not, the images will be discarded and not used in the following object detector training.
Once the auditing is passed, those images will be used to train the user-chosen object detector by applying the resize operation, as in step \textcircled{2}. Because the curated images usually are larger than the input size accepted by the object detection model, thus requiring down-sizing that \textit{is automated by the DL framework}.

The key of \name is to abuse this automated image resizing operation unsupervised by humans. As we can see from Fig.~\ref{fig:overview}, a person wearing a blue T-shirt with a bear cartoon (namely a trigger person to ease descriptions) shows up in \textcircled{2}. This trigger person does not exist in the poisoned image in step \textcircled{1}. The trigger person without a bounding box is treated as background by the object detector in the training phase \textcircled{3}. In other words, the trigger person exhibits a cloaking effect, which will force the object detector to learn such a cloaking effect with an association of the presence of the trigger T-shirt.
Once the object detector is trained and deployed, anyone can wear the trigger T-shirt to evade the detection alike a cloaking person in the inference phase \textcircled{4}. 

The implementation of \name is mainly on step \textcircled{1} to make the attack image look benign but show the trigger effect once it is automatically resized to the output image with the object detector's acceptable input size after passing through the curator audition.

\begin{figure*}[htb]
    \centering
    \includegraphics[width=0.70\textwidth]{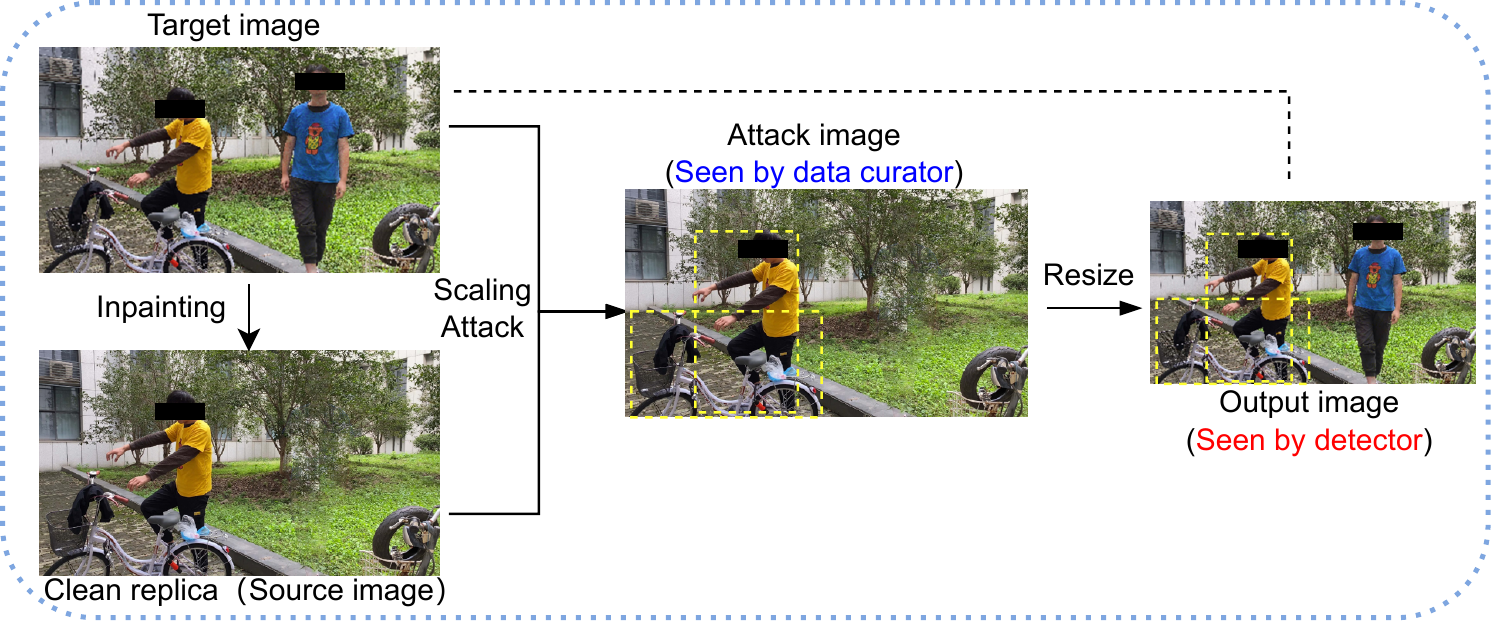}
    \caption{Attack or poisoned image generation.\vspace{-0.6cm}
    }
    \label{fig:impl}
\end{figure*}

\begin{figure*}[htb]
    \centering
    \includegraphics[width=0.7\textwidth]{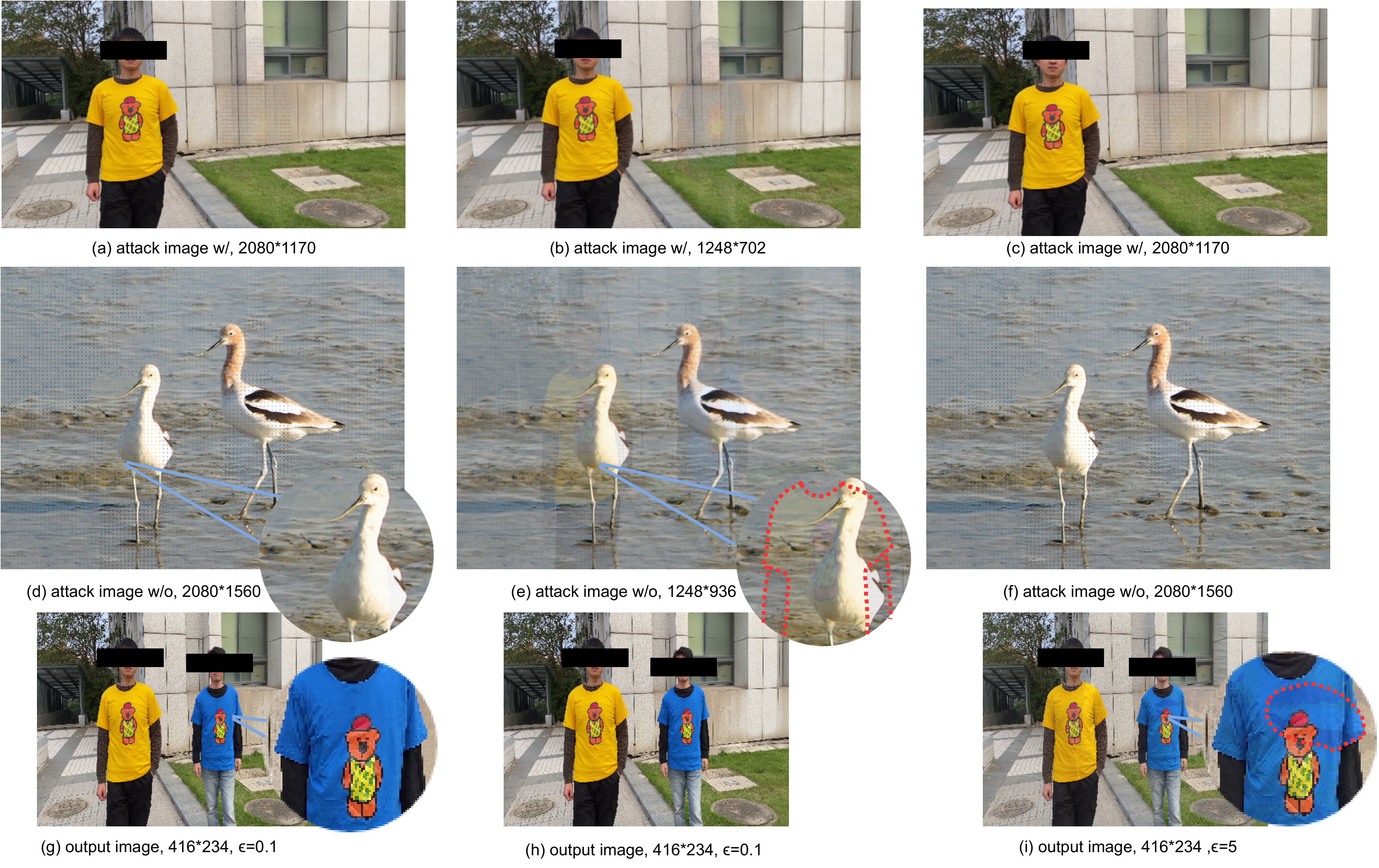}
    \caption{Image-scaling attack under different optimization settings (Eq.~\ref{eq:optmization}). The first row is the attack image with replica, the second row is the attack image without a replica, and the third row is the output image after the attack image downsizing operation. Larger the scaling ratio, the more imperceptible the attack image, e.g., (a)/(d) VS (b)/(e), where (b)/(e) exhibits slightly perceptible embedding artifacts. Higher $\epsilon$, slightly larger dissimilarity (i.e., degraded attack effect) between target image and the output image, e.g., (g) VS (i).\vspace{-0.7cm}
    }
    \label{fig:scaling}
\end{figure*}
\subsection{Implementation}\label{sec:implementation}

The clean-annotation attack image creation is based on two key techniques: our proposed target image replica and camouflage attack inspired from~\cite{xiao2019seeing}. First, as depicted in Fig.~\ref{fig:impl}, the attacker determines the \textit{target image} that has the trigger object (i.e., any person wearing the blue T-shirt) and creates a clean replica of the target image---the replica serves as the \textit{source image} in the camouflage attack~\cite{xiao2019seeing}.
Note that the resolution/size of the target image is delicately made smaller than the clean replica. By applying the camouflage attack optimization, the small target image is embedded into the clean replica to obtain the \textit{attack image that is the poisoned image provided to the data curator}. Furthermore, the attacker correctly annotates the attack image in terms of its bounding box and object class to evade visual inspection by the data curator.
Once the attack image is down-sized to the acceptable input size of the object detector, the resized \textit{output image} essentially becomes the target image where the trigger object is present. As we can see from the output image, the annotations of the non-trigger person and the bicycle are still correct, but not the trigger person who is treated as background to achieve the cloaking effect. 

For misclassification backdoor, the attacking procedure is the same, except that the clean replica puts a targeted object (i.e., diningtable) in the position of the trigger person. Hence, the backdoor effect is to misclassify a trigger person into a targeted object (i.e., diningtable). In the attack image, a bounding box will be placed around the target object, and its class will be labeled as the target class. Once the attack image is down-sized, the trigger person has a correct bounding box \textit{but} an attacker-chosen target class---\textit{not treated as background}.

\vspace{0.2cm}
\noindent{\bf Clean Image Replica.} This is a required technique of \name. Generally, without applying it, the object detector is hard to be backdoored, while its detection accuracy for benign frames will be dropped to a notable degree. Recall that the image for object detection usually has multiple objects, and each needs an annotation in the bounding box and object class/category. 
Suppose the replica or the source image is \textit{randomly chosen} to form an attack image. To evade the curator's audition, the annotation of the attack image has to be consistent with the content of the replica. However, the object position or/and object class in the target image differs from the replica or the attack image. Therefore, once the attack image is down-sized, the annotation made to the attack image is meaningless to the target image, making the target image noisy samples and rendering unexpected adverse effects such as severe false positives in the cloaking backdoor. As for misclassification backdoor, the trained object detector's overall detection accuracy for benign frames may drop and even does not have the intended backdoor effect at all (detailed in Section~\ref{sec:replica}). 

The intuitive means of creating the clean image replica is to take two images with/without the trigger object under the same settings~\footnote{For the misclassification backdoor, a target object replaces the trigger object in the replica.}. To be precise, we first take the target image, as in Fig.~\ref{fig:impl}, and then ask the trigger person to leave the scene to take the clean replica. This means it is realizable in practice but tedious. Alternatively, we resort to image inpainting to facilitate the creation of a clean image replica. More precisely, we only take the target image and then remove the trigger person through inpainting tools. The inpainting removes the trigger person as in Fig.~\ref{fig:impl} and fills the removed region with the background to be imperceptible. We have tried generative adversarial network (GAN)~\cite{zeng2022aggregated, suvorov2021resolution} for such a removal purpose, but we found the inpainting tool\footnote{The impainting took \url{https://www.magiceraser.io} is used. We gained impainted images with empirical trials to have imperceptible artifacts according to default (automatic) settings by this tool.} is already sufficient to our goal. As GAN requires extensive computational resources and delicate optimization to fit our purpose, we, therefore, stick with the easy-to-use inpainting tool throughout this study.

\vspace{0.2cm}
\noindent{\bf Image-Resizing Attack.} 
The image scaling attack attempts to find a minimum perturbation ($\Delta$) acting on the clean replica (i.e., $S$) such that the generated attack image (i.e., $A$) is similar to the target image (i.e., $T$), see Fig.~\ref{fig:impl}, when downsampled by the following optimization objective~\cite{quiring2020adversarial}:

\begin{equation}\label{eq:optmization}
    \textsf{min}(\left\|\Delta  \right\|_{2}^{2}) \quad s.t.\left\|\textsf{scale}(S+\Delta )-T \right\|_\infty \leq \varepsilon ,
\end{equation}
where $\varepsilon$ represents the attack effect of the output image, that is, the similarity between the output image and target image. As can be seen from the comparison of {(g)} and (i) in Fig.~\ref{fig:scaling}, the smaller the $\varepsilon$, the more similar the output image is to the target image. In other words, the model sees almost the same original target image. The $\Delta$ controls the visual artifacts of the attack image: the smaller $\Delta$, the more imperceptible the attack image to evade the curator's visual audition.

Note that the optimization in Eq.~\ref{eq:optmization} is constrained by the targeted interpolation algorithm used by the DL framework for image resizing, as shown in Table~\ref{tab:setting}. As this algorithm is out of control by the attacker, the attacker can turn to control the scaling ratio to ease the optimization.
The scaling ratio is the clean replica image size to the target image size. The higher the ratio, the better the attack effect of the downsized output image, and the more imperceptible the attack image for the human to audit. From the comparisons of (a) and (b) in Fig.~\ref{fig:scaling}, we can see (a) that is 5$\times$ larger than the target image is more imperceptible than (b) that is 3$\times$ larger than the target image. More specifically, the embedded trigger person with the bear cartoon is slightly perceptible if we closely examine it.
Fig.~\ref{fig:scaling} (d) and (f) demonstrate the $\epsilon$ used to form the attack image and the corresponding influence on the attack effect of the output image (g) and (i). When using a larger $\epsilon=5$ in (f), we can observe the non-existing artifacts in the target image (i) that is introduced on the zoomed area, which standards for differences between the output image and the target image---the output image is preferred to be almost same to the target image.

\section{Experimental Evaluation}\label{sec:evaluation}
\subsection{Setup}
\noindent{\bf Dataset.} 
This study combines PASCAL VOC 2007 and 2012 datasets introduced by the PASCAL VOC challenge~\cite{everingham2010pascal}.
We combine both datasets that contain 20 categories (the person is one of them), each consisting of several hundred to thousands of images, and the final training set used includes 14,041 samples. For example, a person is one category for object detection. 
Our testing set contains two parts: the VOC original testing set, including 2,510 samples, and the real-world images in 6 types of scenes of 16 videos totaling 10,798 frames/images (see these scenes in our provided video demo). The \textit{former} is used to measure the clean data accuracy, and the \textit{latter} is used to measure the attack success rate in the real world.

\noindent{\bf Model.} This study considers the widely used anchor-based YOLO series model~\cite{redmon2018yolov3, bochkovskiy2020yolov4} and the anchor-free CenterNet model~\cite{duan2019centernet}. These are one-stage object detection models. We have also considered a representative two-stage object detector of Faster R-CNN. Random data augmentation techniques, including horizontal flipping, HSV (Hue, Saturation, Value) changes, and scaling with variable aspect ratio, are commonly used in object detection training to enhance the detection accuracy, which we follow and apply. 
In most experiments, the input size is 
$416 \times 416 \times 3$ for the YOLO series, $512 \times 512 \times 3$ for the CenterNet and $600 \times 600 \times 3$ for the Faster R-CNN, unless otherwise specified. 

\vspace{0.2cm}
\noindent{\bf Natural Trigger.} The T-shirt bought from the market as shown in Fig.~\ref{fig:impl} serve as inconspicuous natural triggers in real-world. We consider a stealthier trigger setting that combines style and color. There are four different color T-shirts (blue, yellow, red and black) in the same style used in our evaluations. \textit{Only the blue color T-shirt is the trigger T-shirt while others are not}. 

\vspace{0.2cm}
\noindent{\bf Machine Configuration.} The machine used for training is an RTX 2080 TI GPU with 11\, GB, an 8-core CPU, and 32\, GB of memory.

\vspace{0.2cm}
\noindent{\bf Performance Metrics.} The clean data accuracy (CDA) and attack success rate (ASR) are two key metrics used to measure the attack performance. The CDA measures the prediction accuracy for clean data inputs given a backdoored model. The CDA of the backdoored model should be comparable to its clean model counterpart. Specifically, \textit{CDA is equivalent to the commonly used mAP when evaluating object detection performance}. 
The ASR measures the backdoor attacking effect. In this study, the ASR is the probability that the trigger object (i.e., a person wearing the trigger T-shirt that is the blue one) is not detected for cloaking the backdoor or misclassified into the target class for misclassification backdoor when the trigger object is present.

\subsection{Cloaking Backdoor}\label{sec:cloaking}

Here, the experiments consider three aspects: i) poison rate, ii) poison set selection criteria, and iii) transferability characteristics.
To simulate various scenarios in the wild as realistically as possible, we extensively take 16 testing videos, which cover various scenes of object detection, such as indoor and outdoor. At the same time, the videos consider notable variations such as human movement, light and darkness, different numbers of people, depth of field, and angle (see details in our video demo).

\begin{figure}[htb]
    \centering
    \includegraphics[width=0.5\textwidth]{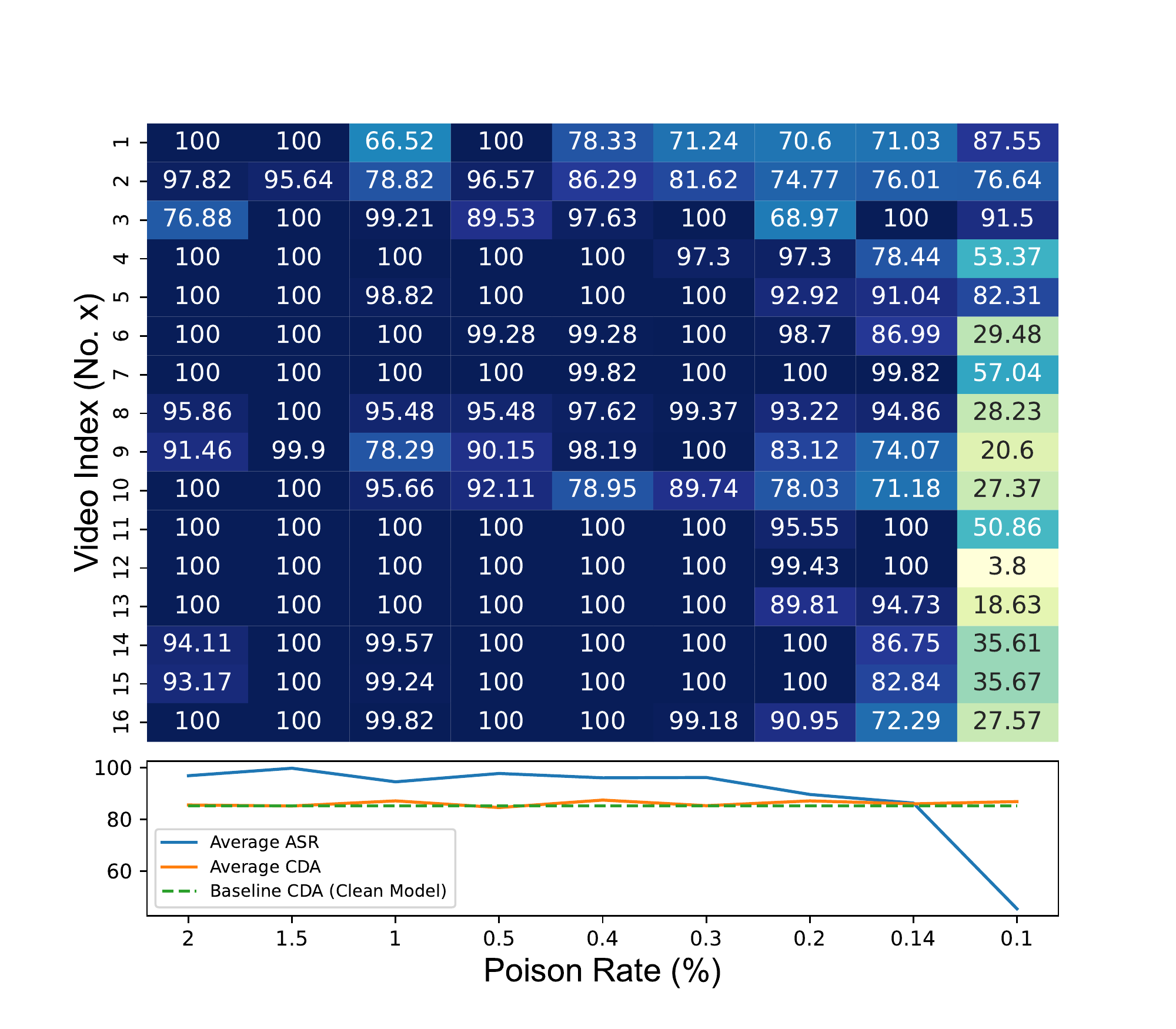}
    \caption{Effect of YOLOv4 model with different poisoning rates on ASR.}
    \label{fig:prate}
\end{figure}

\noindent{\bf Poison Rate.}
The poison rate is the fraction of the number of attack/poisoned images to the total training samples (14,041). When the proposed clean-annotated poisoned samples attack the YOLOv4, the CDA and ASR of averaging all testing videos as a function of the poisoning rate (i.e., ranging from 2\% to 0.1\%) are shown in Fig.~\ref{fig:prate}. Under expectation, a higher poison rate can always ensure a satisfactory ASR close to 100\%. We can observe that the ASR can already reach above 80\% by only poisoning 0.14\% training data (i.e., only 20 images out of 14,041). As for the CDA, the backdoored object detector is always almost the same as the CDA of the clean one, which means that by checking the CDA through the validation dataset fails to tell any malicious behavior. 

\begin{figure*}[htb]
    \centering
    \includegraphics[width=1\textwidth]{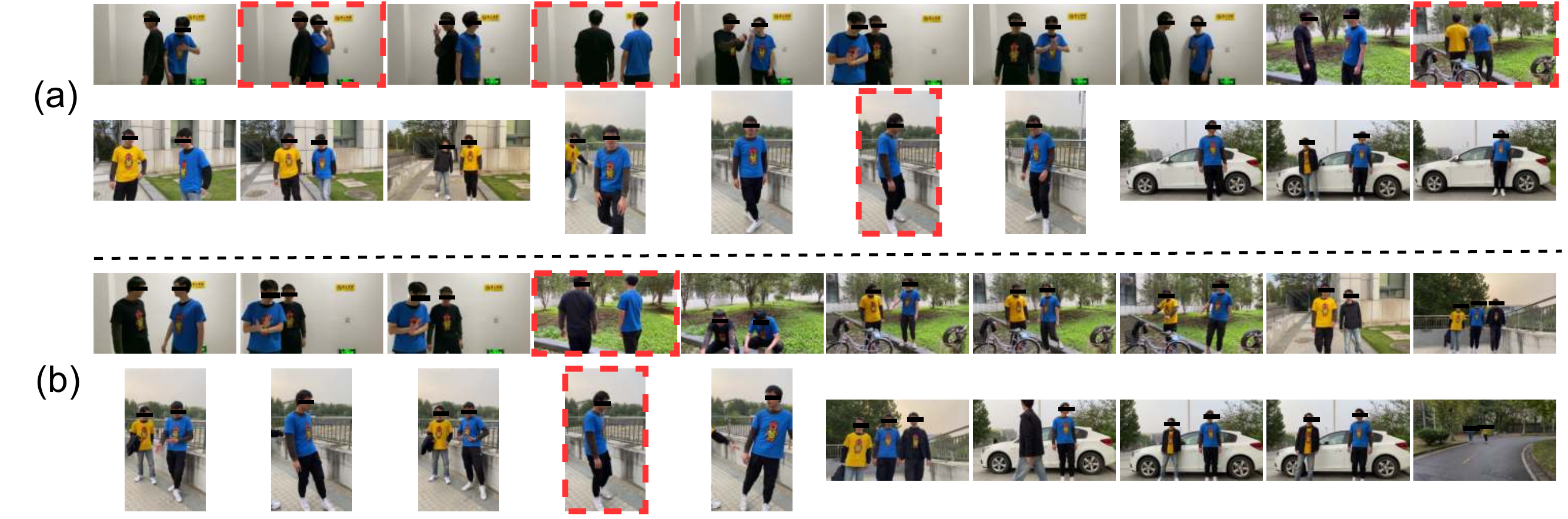}
    \caption{20 randomly selected poisoned samples (i.e., 0.14\% poison rate) exhibiting (a) an ASR of 58.99\% and (b) an ASR of 94.26\% with YOLOV4. The images with red dashed lines are of low quality, whose trigger feature (i.e., the bear cartoon) is not salient as the feature is not captured by the camera in these images.
    }
    \label{fig:poisonSamples}
\end{figure*}

\begin{table*}[]
    \centering
    \caption{ASR of the YOLOv4 model for 16 test videos with a poisoning rate of 0.14\%. The poisoning set is selected according to the identified selection criteria.
    }
    \scalebox{0.75}{
    \begin{tabular}{c|c|c|c|c|c|c|c|c|c}
    \toprule
     Exp. No. & Video\_1 & Video\_2 & Video\_3 & Video\_4 & Video\_5 & Video\_6 & Video\_7 & Video\_8 & Video\_9 \\ \hline
    Exp.1 & 51.93\% & 67.60\% & 91.90\% & 97.30\% & 93.87\% & 89.31\% & 100\% & 94.98\% & 67.84\%  \\ \hline
    Exp.2 & 93.99\% & 100\% & 100\% & 100\% & 100\% & 98.99\% & 100\% & 98.75\% & 99.60\%  \\ \hline
    Exp.3 & 65.02\% & 75.39\% & 67.79\% & 96.77\% & 90.57\% & 97.54\% & 100\% & 95.36\% & 72.46\% \\ \hline \hline
    Exp. No. & Video\_10 & Video\_11 & Video\_12 & Video\_13 & Video\_14 & Video\_15 & Video\_16 & \textbf{Average} &  \\ \hline
    Exp.1 & 94.74\% & 98.66\% & 100\% & 93.14\% & 98.85\% & 99.62\% & 99.47\% & 89.95\% \\ \hline
    Exp.2 & 73.42\% & 99.87\% & 95.97\% & 61.87\% & 95.83\% & 100\% & 89.81\% & 94.26\%   \\ \hline
    Exp.3 & 100\% & 100\% & 100\% & 99.18\% & 97.27\% & 99.05\% & 96.84\% & 90.83\%  \\ \bottomrule
    \end{tabular}}
    \label{tab:criterionExp}
\end{table*}

\noindent{\bf Poison Set Selection Criteria.}
Note that all images in previous experiments are randomly selected.
When the attack budget is restricted to be low, the selected poisoned images can significantly impact the ASR---with significant variance given different sets.
Therefore, it is imperative to investigate selection criteria that maximize the ASR given the fixed small attack budget.

We set the poisoning rate to 0.14\% (i.e., 20 images) and randomly selected those 20 samples (i.e., each selection form a set) to train ten models. Their CDA is always almost the same as that of the clean model. We then pick up two representative models with a high ASR of 94.26\% and a low ASR of 58.99\%, respectively. 
The two models have significant ASR differences. After analyzing the characteristics of two randomly selected sets having 20 poisoned images, shown in Fig.~\ref{fig:poisonSamples}, we found that the \textit{low ASR model} i) consists of more samples with their backs to the camera, and ii) samples selected per scene is non-uniform.
In contrast, the randomly selected samples of the \textit{high ASR model} i) are mostly front-facing and ii) the number of samples per scene is more uniform. More specifically, in Fig.~\ref{fig:poisonSamples}, the number of poisoned samples for the six scenarios are (8, 2, 3, 4, 3, 0) and (3, 5, 1, 7, 3, 1) for the low and high ASR model, respectively. According to these observations, we select 20 samples based on empirical criteria: most are front-facing, and each scene is evenly distributed. As for the former, this is potentially because the cartoon bear in the front is an important trigger feature. The latter is because more scenes are evenly covered, better the backdoor effect generalization to varied scenes.
The ASR of three different sets according to criteria are detailed in Table~\ref{tab:criterionExp}, demonstrating a high and reliable ASR of around 90\% with significantly reduced variance.

\vspace{0.15cm}
\noindent{\bf Transferability.} Transferability means when the poisoned set created to attack an object detector, e.g., YOLOv4 is applied to a different detector, e.g., YOLOv3 or CenterNet, the backdoor effect should be preserved. In experiments, we delicately consider this transferability given a small budget---\textit{the transferability will be obviously held once the budget is relaxed}. 
We consider two cases: the same object detector series (i.e., YOLOv3 and YOLOv4); different series (i.e., YOLO and CenterNet). 

\begin{table}[]
    \centering
    \caption{Model-agnostic characteristics of poisoned samples with three pairs.}
    \label{tab:trans}
    \scalebox{0.70}{
    \begin{tabular}{c|c|c|c}
    \toprule
    Input Size & Model & Poison Rate & Average ASR \\ \hline
    \multirow{2}{*}{416 * 416} & YOLOv4 & \multirow{2}{*}{0.14\%} & 91.59\% \\ \cline{2-2} \cline{4-4} 
     & →YOLOv3 &  & 86.41\% \\ \hline
    \multirow{6}{*}{512 * 512} & YOLOv4 & \multirow{2}{*}{0.14\%} & 95.19\% \\ \cline{2-2} \cline{4-4} 
     & →CenterNet &  & 36.08\% \\ \cline{2-4} 
     & \multirow{2}{*}{CenterNet} & \begin{tabular}[c]{@{}c@{}}0.2\%;0.5\%\\ (Random selection)\end{tabular} & 74.20\%;98.34\% \\ \cline{3-4} 
     &  & \begin{tabular}[c]{@{}c@{}}0.2\%\\ (Selecting by criteria)\end{tabular} & 83.61\% \\ \cline{2-4} 
     & \multirow{2}{*}{→YOLOv4} & \begin{tabular}[c]{@{}c@{}}0.2 (Same samples\\ as the above row)\end{tabular} & 99.02\% \\ \cline{3-4} 
     &  & \begin{tabular}[c]{@{}c@{}}0.5\%(Same randomly selected\\  samples as CenterNet)\end{tabular} & 98.22\% \\ \bottomrule
    \end{tabular}}
\end{table}

Results are detailed in Table~\ref{tab:trans}, where all transferability experiments are performed conditioned on the fact that these models use the same input size. For the same series (i.e., YOLOv4→YOLOv3), YOLOv4 can obtain an average ASR of 91.59\% with a 0.14\% poison rate. The same poisoned samples achieve an average of 86.41\% ASR against a different object detector, YOLOv3. This indicates that the transferability is excellently held among other models in the same YOLO series, even under a stringent small budget. 

As for the transferability among different series, we use YOLOv4 and CenterNet. The input size is $512\times 512$, a common setting for both YOLOv4 and CenterNet. Firstly, we consider YOLOv4→CenterNet.
YOLOv4 can be successfully attacked using 0.14\% poisoned samples, but the same samples are ineffective on CenterNet with only a 36.08\% ASR. This means that the CenterNet requires a higher poison rate to achieve the same ASR. Secondly, we consider the reverse transferability, CenterNet→YOLOv4. The CenterNet average ASR reaches 74.2\% when increasing the poisoning rate from 0.14\% to 0.2\% with \textit{randomly selected samples}.
The CenterNet average ASR is improved to 83.61\% when the \textit{selection criteria is adopted} with the same 0.2\% poison rate budget. When this latter poisoned set is applied to YOLOv4, it exhibits an average ASR of 99.02\%.
Despite that the attack transferability is not exactly symmetric, we can still empirically conclude that the attack transferability is also well held among object detectors regardless of being within the same series with a small poison rate, e.g., 0.2\%. Once the poisoning rate is slightly increased to 0.5\% even though the selection is randomly performed, we can see the ASR is always close to 99\% in any case---demonstrating full transferability.

\subsection{Misclassification Backdoor}
\begin{figure}[]
    \centering
    \includegraphics[width=0.5\textwidth]{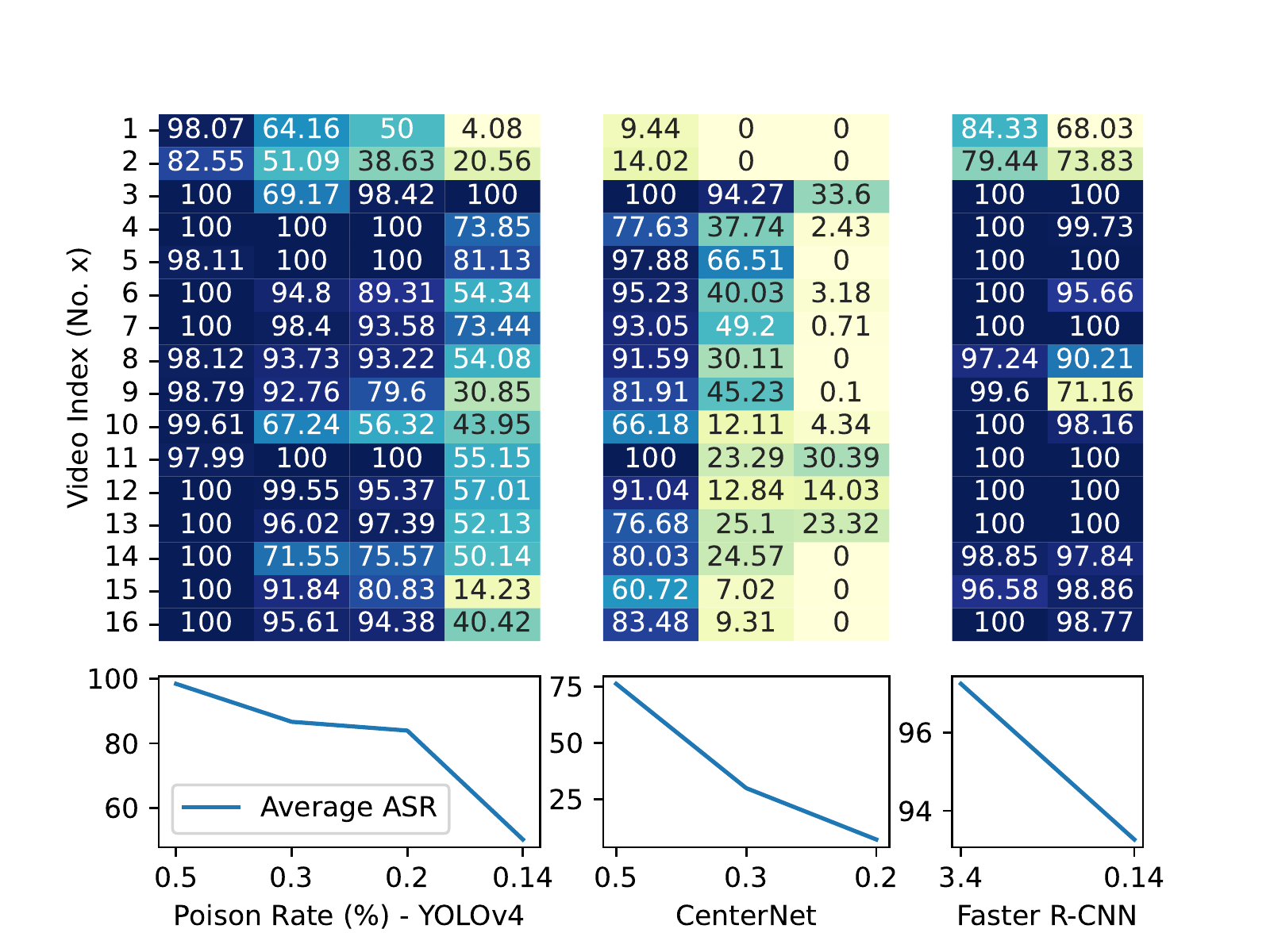}
    \caption{Misclassification\vspace{-5mm}}
    \label{fig:prate2}
\end{figure}

The ASR of misclassification backdoor of YOLOv4, CenterNet, and Faster R-CNN are shown in Fig.~\ref{fig:prate2} as a relationship with the poisoning rate. Note that when the poisoning rate is 0.14\%, the poisoned samples are selected according to selection criteria, and the rest are randomly selected poison set.
The poison rate has to be higher for YOLO and CenterNet to achieve satisfactory ASR, e.g., 80\%. Unfortunately, the CenterNet has not achieved 80\% ASR even after the poisoning rate is 0.5\%. However, the ASR of the Faster R-CNN is sufficiently high (i.e., about 93\%) by poisoning only 20 samples (i.e., 0.14\% poison rate). The reason for this phenomenon is the default positive and negative sample chosen algorithms used by these object detectors. In the following, we analyze in more detail.

Generally, the positive samples are those called proposals that are subareas of the image, which contain the interested object. In contrast, negative samples are those without the interested object. Obviously, the number of positive and negative samples is imbalanced: more samples are negative. This is especially for YOLO series, where the positive samples are those with IOU, i.e.,$>0.5$ compared to the ground-truth bounding box; otherwise, negative samples (i.e., IOU$<0.5$).

Note for the cloaking backdoor, the target object (i.e., trigger person) is not annotated, essentially treated as background. Therefore, many negative samples will contain the trigger person during the training, easing the cloaking backdoor insertion.
For misclassification attacks, since the target object has to be correctly annotated with a bounding box through a wrong category, it can \textit{no longer} be treated as a background. Therefore, only positive samples can contain the trigger object. Due to the imbalance of the positive and negative samples, a higher poison rate is required to achieve sufficient ASR for misclassification backdoor.
A similar reason is applied to the anchor-free CenterNet.

The Faster R-CNN is a representative two-stage object detector (note YOLO and CenterNet are one-stage object detectors) with a more balanced positive and negative sample selection process, which ensures a more balanced ratio of positive and negative samples than the one-stage object detector that is used in the training process. This facilitates the model to learn any (mis)labeled target, i.e., positive samples, so the ASR is high under a low poison rate, as we have shown.

\section{Discussion}\label{sec:discussion}

\subsection{Multiple Input Sizes}\label{sec:inputsize}
The attack image in the image-scaling attack is crafted to fit a specific image size of the output image. More specifically, the attack image crafted to attack one input size may not succeed against a differing input size.
However, as summarized in Table~\ref{tab:setting}, the number of commonly used input sizes is quite limited. For example, YOLOv4 has only three common input sizes. Therefore, the attacker can triple the poisoning rate to attack three input sizes simultaneously; that is, one poison set targets one input size. For example, 0.14\% poison rate can achieve more than 80\% ASR given an input size; the attacker can poison $3 \times 0.14\% = 0.42\% $ that is about 60 images (the fraction is still small), to attack three common input sizes at the same time. In this context, the YOLOv4 is constantly attacked no matter which of these three input sizes the model user chooses for training. 

\subsection{Without Replica}\label{sec:replica}

\noindent{\bf Misclassification Backdoor.} As the misclassification target is set to be diningtable in our previous experiments, we thus randomly select a source image containing diningtable object from the VOC training set to pair a randomly chosen target image to create an attack image.
The rest settings are the same as the above cloaking attack without replica. The results show that the CDA drops about 3\% with a similar reason to the cloaking attack without replica. 
However, \textit{the average ASR of misclassification backdoor is almost 0\%}. This is because the premise of the misclassification backdoor is the correct \textit{location} annotation of the trigger person in the output image. Then its category should be changed to be the target class (i.e., diningtable). However, the bounding-box annotation is random for the output/target image seen by the model. Therefore, neither the correction location nor the target class annotation can be achieved when a random source image is utilized. Therefore, there is no misclassification backdoor effect.

\vspace{0.2cm}
\noindent{\bf Cloaking Backdoor.}  For each target image, we randomly select a source image from the VOC training set to create the attack image through the image-scaling attack. Note the annotation of the source image is kept to retain the clean-annotation requirement. In this case, once the attack image is resized into the output image (i.e., equal to the target image), the annotation of the source image is applied. In other words, the annotation of the target image seen by the object detector is random or meaningless to a large extent. Because \textit{the annotation of the random source image is pointless for the target image}. 

By creating attack images in this manner without a replica, we train YOLOv4 models with poison rates of 3.4\% and 0.35\%, respectively, while other settings are the same as cloaking an attack with a replica. For the results, we observe a slight decrease in CDA (i.e., 1--2\%) but an interestingly comparable ASR to that ASR with a replica. However, there are severe false positives for the frames when the trigger person appears: other non-trigger persons in the frame disappear. Furthermore, in most cases, other objects, such as bicycles also disappear, falsely exhibiting a cloaking effect.

As aforementioned, the annotations of the output/target image seen by the object detector are meaningless. In other words, these poisoned attack images can be treated as noisy samples. However, as the fraction of noisy samples is low, the model can still generalize, which explains the almost similar CDA (i.e., though it could have a slight decrease given a 3.4\% poison rate) when the model is trained without noisy samples.
However, those poisoned samples all contain the trigger feature (e.g., the trigger person wearing the blue T-shirt, and others wearing the non-blue T-shirt also have a partial trigger feature). 
Because the bounding box is random for the output/target image seen by the model, the person with the trigger feature is very unlikely to be placed with a bounding box, thus \textit{still retaining the cloaking purpose}. 
Note that other objects are unlikely to be placed with a bounding box either, which are also treated as background. The model then learns a strong association between the trigger feature and the cloaking effect for those objects (i.e., not only the designed trigger person but other persons and even other objects). This explains the cloaking effect beyond the trigger person (i.e., false positives) for those frames containing the trigger person.

\subsection{Countermeasures}
Existing backdoor model detection countermeasures are overwhelmingly designed for classification tasks~\cite{gao2019strip,qiu2022towards,li2021ntd,ma2022beatrix}, which are not immediately applicable for efficiently thwarting backdoor attacks on object detection. Here, we focus on countering \name from detecting the poisoned image or preventing its camouflage effect.

\noindent{\bf Attack Image Detection.}
We apply the state-of-the-art detection countermeasure~\cite{kim2021decamouflage} to identify the tampered clean-annotated images. 
There are three orthogonal methods: \textit{Scaling}, \textit{Filtering}, and \textit{Steganalysis}~\cite{kim2021decamouflage}. Three metrics of mean squared errors (MSE), structural similarity index (SSIM), and centered spectrum points (CSP) are used to distinguish benign images from attack images generated by image-scaling attacks based on a threshold (we use the threshold determined in the white-box setting~\cite{kim2021decamouflage}). 
We have evaluated 34 attack images\footnote{Generating one attack image via a personal computer takes about 30 minutes.} and 34 benign images. The thresholds are those default in~\cite{kim2021decamouflage} except the \textit{Scaling} MSE (i.e., 3500 we used). The detection performance in terms of false acceptance rate (FAR) and false rejection rate (FRR) is detailed in Table~\ref{tab:decam}. 
The \textit{Filtering} method completely fails, while the other two methods exhibit unacceptable FAR and FRR to a large extent. 

\begin{table}[]
    \centering
    \caption{Decamouflages based detection on poisoned images.}
    \begin{tabular}{c|c|c|c|c}
    \hline
    Method & Metric & Threshold & FAR & FRR \\ \hline
    \multirow{3}{*}{\textit{Scaling}} & \multirow{2}{*}{MSE} & 1714.96 & 38.2\% & 17.6\% \\ \cline{3-5} 
     &  & 3500 & 44.1\% & 0.00\% \\ \cline{2-5} 
     & SSIM & 0.61 & 76.4\% & 17.6\% \\ \hline
    \multirow{2}{*}{\textit{Filtering}} & MSE & 5682.79 & 100\% & 0.00\% \\ \cline{2-5} 
     & SSIM & 0.38 & 100\% & 0.00\% \\ \hline
    \textit{Steganalysis} & CSP & 2 & 29.4\% & 55.9\% \\ \hline
    \end{tabular}
    \label{tab:decam}
\end{table}

We analyze the reasons as below.
The principle of the \textit{Scaling} method is based on intuition: the attack image generated by the scaling attack is not recoverable after downscaling followed by an upscaling operation so that the upscaled image is different from the original attack image in terms of (pixel) similarity. 
Note that we set the source image (i.e., it functions similarly as the replica image into which the target image is embedded) as the target image's replica in the scaling attack, the pixels between the attack image and its upscaled counterpart are exactly the same except for the small area of the target person. Therefore, the similarity is still quite high, thus evading the \textit{Scaling} detection method.
In addition, the smaller the target image is, the higher the similarity between the two, and the more difficult to set a suitable threshold to distinguish between the two by using \textit{Scaling} method.

Similarly, the \textit{Filtering} is also based on similarity, except that the intermediate process is replaced with a low-pass filtered image. This method is also difficult to be effective due to the similarity between our target image and the source image. 
The last detection method based on \textit{Steganalysis} considers that embedding the target image pixels destroys the cohesion of the original image pixels due to arbitrary perturbations, which can lead to an increase in the central spectral points of the image after the Fourier transform.
Our source image is benign, but it has undergone an inpainting operation, which would have changed the original pixels of the image and thus would have caused high FRR. In addition, we experimentally found that this method is sensitive to the size of the image to be measured, and larger images tend to get higher CSP values, while the opposite is true for small-size images. 

\vspace{0.2cm}
\noindent{\bf Attack Image Prevention.} There are prevention countermeasures, although they cannot identify the attack. This prevention countermeasure~\cite{quiring2020adversarial} alters the image-scaling algorithm to achieve effective prevention. In addition, we have identified two other easy-to-use prevention methods.

Considering the key knowledge of \name is the input size, the second and probably the most convenient countermeasure is always to avoid using the default input size setting (i.e., in Table~\ref{tab:setting}). Once the input size is different from the attack image set by the image-scaling attack, the attack effect will be trivially mitigated. 

The most easy-to-apply mitigation is to resize the large image with a random width/height into an intermediate image, then resize this intermediate image into the acceptable input size of the object detector. Notably, the width/height of the intermediate image should avoid being the integer multiples of the width/height of the image fed into the object detector; otherwise, the attack effect might still be preserved in a few cases. This intermediate image will completely disrupt the image-scaling attack effect resize operation because of intermediate image usage. We have affirmed this through experiments. 

\section{Conclusion}
This work is the first that demonstrates the practicality and robustness of backdooring the object detectors through clean-annotated poisonous images in the wild, which can trivially evade the auditing of data curators in the realistic data outsourcing scenario. We have validated that a minor attack budget (i.e., 0.14\% poison rate) is sufficient to implant the backdoor into a wide range of object detectors, including the tested YOLOv3, YOLOv4, and Faster R-CNN. Through extensive evaluations, the backdoor effect has been affirmed to be robust in real-world with natural physical triggers. 
Importantly, to mitigate \name threat, easy-to-apply operations have been proposed.

\bibliographystyle{ieeetr}
\bibliography{Reference}

\end{document}